\documentclass{article} \usepackage{iclr2027_conference,times}

\usepackage{amsmath,amsfonts,bm}

\def\eqref#1{equation~\ref{#1}}

\def\1{\bm{1}}

\DeclareMathAlphabet{\mathsfit}{\encodingdefault}{\sfdefault}{m}{sl}
\SetMathAlphabet{\mathsfit}{bold}{\encodingdefault}{\sfdefault}{bx}{n}

\usepackage{hyperref}
\usepackage{url}
\usepackage{amsmath}
\usepackage{amssymb}
\usepackage{xcolor}
\usepackage{algorithm}
\usepackage{algpseudocode}
\usepackage{diagbox}
\usepackage{booktabs}
\usepackage{multirow}
\usepackage{enumitem}
\usepackage{subcaption}
\usepackage{wrapfig}
\usepackage{graphicx}

\title{REASON FIRST, RETRIEVE WHAT MATTERS: HYPOTHESIS-GUIDED KEYFRAME SELECTION FOR LONG VIDEO UNDERSTANDING}

\author{Yisong Hu, Junbo Wang, Yuke Li, and Jiangbin Zheng\thanks{All authors are with the School of Software, Northwestern Polytechnical University, Xi'an 710072, China (e-mail: 2021302831@mail.nwpu.edu.cn; jbwang@nwpu.edu.cn; liyuke@nwpu.edu.cn; zhengjb@nwpu.edu.cn).}\thanks{Corresponding author: Jiangbin Zheng.}}
\iclrfinalcopy
\AtBeginDocument{\fancyhead{}\renewcommand{\headrulewidth}{0pt}\hypersetup{hidelinks}}

\begin{document}

\maketitle

\begin{abstract}
Long video understanding demands that models recover and reason over sparse, temporally distributed evidence within limited frame budgets. Existing keyframe selection (KFS) methods usually retrieve query-relevant frames
from a densely sampled candidate set before inference. However, these methods still have two key limitations. First, the global
image--text matching strategy introduces unnecessary computation overhead when a small set of
uniformly sampled frames already provides sufficient evidence. Moreover, they may miss evidence that distinguishes competing answers, while
single-pass frame selection provides no opportunity to correct selection errors.
To address the issues, we propose \textbf{HypoKFS}, a training-free and plug-and-play framework
that couples keyframe selection with adaptive multi-step inference. Specifically‌, HypoKFS first
reasons over sparsely sampled frames to construct an initial hypothesis space. Then the resulting answer distribution determines whether further frame
selection is required. When additional frame selection is needed, the retained hypotheses guide
discriminative frame selection, followed by renewed inference and hypothesis
refinement. This iterative process focuses additional visual computation on
unresolved hypotheses while avoiding unnecessary global frame matching.
Experimental results on the Video-MME, LongVideoBench, and MLVU demonstrate consistent
improvements across different MLLM backbones. HypoKFS improves accuracy by up to 6.7 percentage points and achieves
72.9\% accuracy on Video-MME while applying image--text matching to only
1.3\% of the video frames.
\end{abstract}

\section{Introduction}

Multimodal Large Language Models (MLLMs) have substantially advanced video
understanding and reasoning over temporally structured visual
content~\cite{zhang2024llavavideo,bai2025qwen3vl}. As these models are
extended from short clips to long videos~\cite{fu2025videomme,
wu2024longvideobench,zhou2024mlvu,yang2025keyevl15}, the amount of visual
content grows rapidly while task-relevant evidence often remains sparse over
long temporal contexts~\cite{wu2024longvideobench,tang2025adaptive,
liu2025bolt}. Selecting informative frames under a limited visual budget has
therefore become an important challenge for long-video
understanding~\cite{tang2025adaptive,zhang2025qframe,zhu2026focus}.

\begin{figure}[t]
\centering
\includegraphics[width=0.95\linewidth]{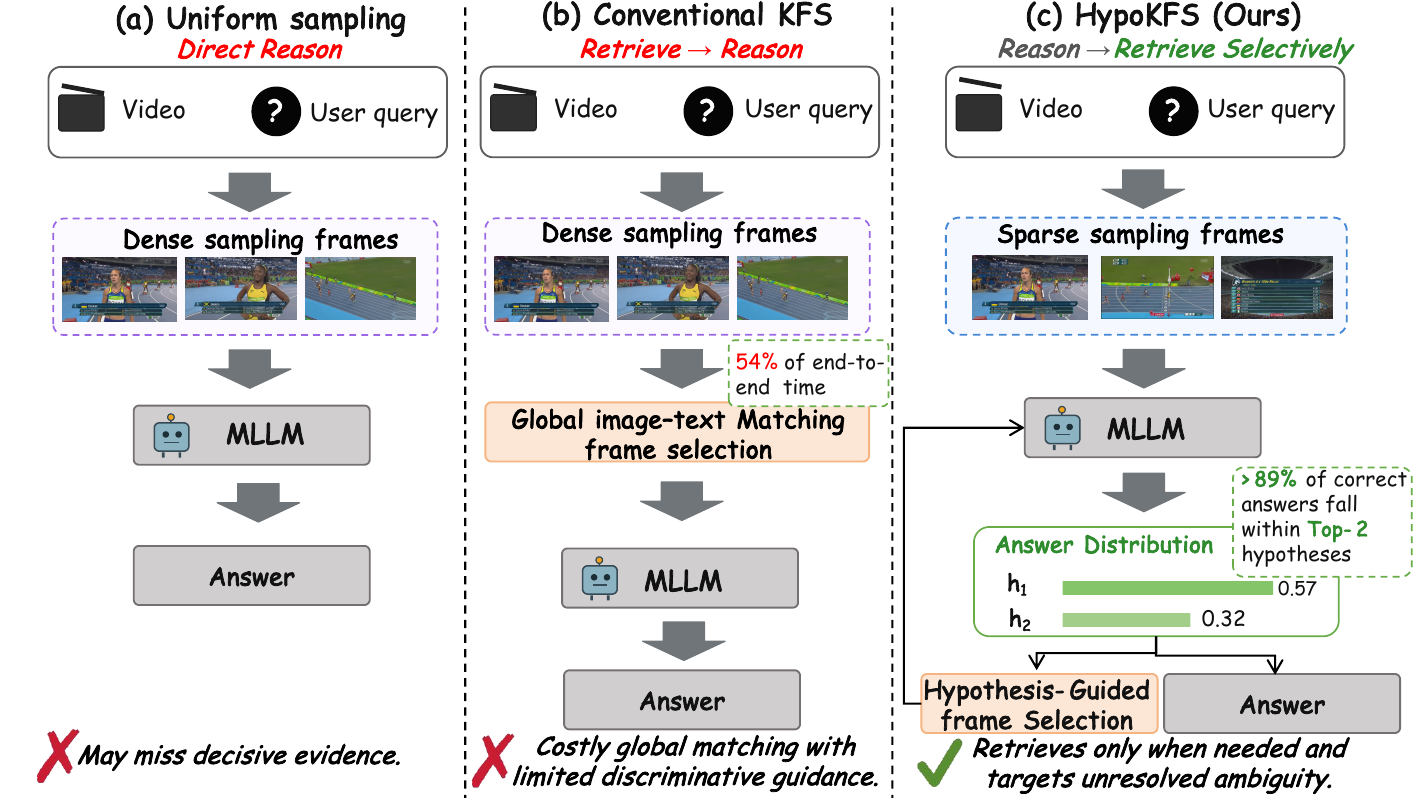}
\caption{
\textbf{Evolution of visual evidence selection for VideoQA.}
\textbf{(a)} Uniform sampling uses fixed, query-agnostic frames.
\textbf{(b)} Conventional KFS performs global image--text matching and
frame selection before MLLM inference.
\textbf{(c)} HypoKFS first infers from sparsely sampled frames and uses
the resulting hypothesis space to guide subsequent frame selection.
}
\label{fig:answer_distribution}
\end{figure}

A common strategy is \emph{uniform sampling}~\cite{fu2025videomme,
wu2024longvideobench,liu2025bolt}, as illustrated in
Fig.~\ref{fig:answer_distribution}(a), where a fixed number of frames are
uniformly sampled before MLLM inference. Despite its simplicity, uniform
sampling is query-agnostic and may miss task-relevant evidence under a
limited visual budget~\cite{liu2025bolt,tang2025adaptive}. Recent keyframe
selection (KFS) methods therefore introduce more targeted frame-selection
strategies~\cite{tang2025adaptive,zhang2025qframe,sun2025mdp3,
zhu2026focus}. As illustrated in Fig.~\ref{fig:answer_distribution}(b),
these methods typically construct a denser candidate set and select
informative frames before MLLM inference using signals such as query
relevance, temporal coverage, diversity, or adaptive exploration.

Despite this progress, performing keyframe selection entirely before
inference does not exploit the predictive information already available from
the MLLM. For some queries, inference over only a small number of uniformly
sampled frames already assigns high confidence to one answer, making
additional global image--text matching unnecessary. For more ambiguous
queries, multiple hypotheses remain plausible and further visual evidence is
required. In such cases, query relevance alone is insufficient for effective
frame selection, since a highly relevant frame may provide similar support
for competing hypotheses and therefore offer limited discriminative value.
Moreover, most existing KFS methods perform frame selection only once before
MLLM inference. If informative evidence is missed during this stage, the
selection cannot be revised according to subsequent model predictions.
These limitations motivate coupling keyframe selection with the evolving
answer distribution rather than treating it as a fixed preprocessing stage.

Fig.~\ref{fig:answer_distribution}(c) provides two observations supporting
this perspective. First, with only uniformly sampled sparse frames, 89.4\% of
correct answers on Video-MME are already covered by the top-2 hypotheses
produced by initial MLLM inference. This indicates that inexpensive initial
inference can provide a compact and informative hypothesis space for
subsequent frame selection. Second, global image--text matching in
conventional KFS accounts for approximately 54\% of the end-to-end inference
time, making unconditional matching over a dense candidate set costly.
Together, these observations suggest that frame matching need not be
performed before inference for every query. Instead, the model can first
estimate the current answer distribution and invoke additional frame
selection only when further visual evidence is required.

Motivated by these observations, we propose \textbf{HypoKFS}, a
training-free and plug-and-play framework for hypothesis-guided iterative
keyframe selection. HypoKFS first performs low-cost inference over uniformly
sampled sparse frames to estimate the current answer distribution and
construct a compact hypothesis space. Confident predictions are returned
directly without additional frame matching, while ambiguous cases trigger
hypothesis-guided frame selection that emphasizes both hypothesis relevance
and discriminative utility. The selected frames are then incorporated into
renewed inference, allowing the updated hypothesis space to guide subsequent
frame selection. This iterative process progressively resolves competing
hypotheses and terminates after a bounded number of rounds, with the
highest-probability hypothesis returned as the final answer.

Our contributions are summarized as follows:
\begin{itemize}[leftmargin=*, nosep]

\item \textbf{Resolving answer ambiguity through iterative frame selection.}
We propose \textbf{HypoKFS}, a training-free and plug-and-play framework
that iteratively selects discriminative frames according to the current
hypothesis space, allowing competing answers to be progressively resolved
while avoiding unnecessary frame matching for confident predictions.

\item \textbf{Reducing redundant computation in keyframe selection.}
We reformulate conventional single-stage keyframe selection as a
hypothesis-guided iterative process, substantially reducing unnecessary
global frame matching and providing a new perspective for efficient
long-video understanding.

\item \textbf{Consistent gains with lower matching cost.}
Experiments on Video-MME, LongVideoBench, and MLVU demonstrate consistent
improvements across different MLLM backbones while substantially reducing
frame--hypothesis matching computation.

\end{itemize}

\section{Related Work}

\subsection{Long Video Understanding}

Long video understanding introduces additional challenges due to the large
number of visual tokens and long-range temporal dependencies. Compared with
short-video settings, long-form videos require models to reason over extended
temporal contexts, substantially increasing computational and memory costs.
Existing approaches mainly address this problem through memory mechanisms,
visual token compression, or long-context modeling.
For example, MovieChat~\cite{2023MovieChat} employs memory banks to maintain
long-term context, while Video-Streaming~\cite{qian2024streaming} and
MA-LMM~\cite{he2024ma} compress video features through compact temporal
representations. LongVLM~\cite{2025LongVLM} and
LongVILA~\cite{2024LongVILA} further extend the temporal context that can be
processed by video MLLMs. More recent methods increasingly focus on reducing
visual-token redundancy. LongVU~\cite{shen2025longvu} performs adaptive
spatiotemporal compression using inter-frame redundancy and cross-modal
signals, while DyCoke~\cite{tao2025dycoke} dynamically compresses visual
tokens and the corresponding KV cache during inference.
PVC~\cite{yang2025pvc} progressively compresses visual tokens across frames,
and VoCo-LLaMA~\cite{ye2025voco} learns compact visual representations that
better align with the LLM's visual understanding process.
Although these approaches substantially improve the scalability of video
MLLMs, they mainly optimize the representation of already sampled visual
content. HypoKFS instead focuses on determining which additional frames
should be processed according to the model's current inference state.

\subsection{Frame Selection}

Frame selection reduces temporal redundancy by retaining a compact set of
informative frames before video reasoning. Early methods such as
AdaFrame~\cite{wu2019adaframe} learn adaptive sampling policies based on
video content, while recent Video-LLM approaches increasingly incorporate
query information into the selection process. Existing methods can be
broadly divided into learning-based and training-free approaches.
FrameVoyager~\cite{yu2024frame} and the M-LLM-based selector of
Hu et al.~\cite{hu2025mllmframesel} learn frame-selection policies from task
supervision or importance signals. In contrast, training-free methods such
as BOLT~\cite{liu2025bolt}, AKS~\cite{tang2025adaptive}, and
Q-Frame~\cite{zhang2025qframe} select frames using query relevance, model
reasoning, or adaptive visual representations without additional training.
Recent approaches further explore complementary selection criteria:
MaxInfo~\cite{li2026maxinfo} selects representative frames by maximizing
diversity in the embedding space, DIG~\cite{li2026dig} adapts the selection
strategy according to query type, and DAFS~\cite{wang2026dafs} derives
frame-level relevance from cross-modal attention at test time. Despite these
advances, most existing methods determine the selected frames before the
final answer inference. HypoKFS instead couples frame selection with the
evolving answer distribution, allowing subsequent selections to be guided by
the unresolved hypothesis space.

\section{Method}

\begin{figure*}[t]
    \centering
    \includegraphics[width=\linewidth]{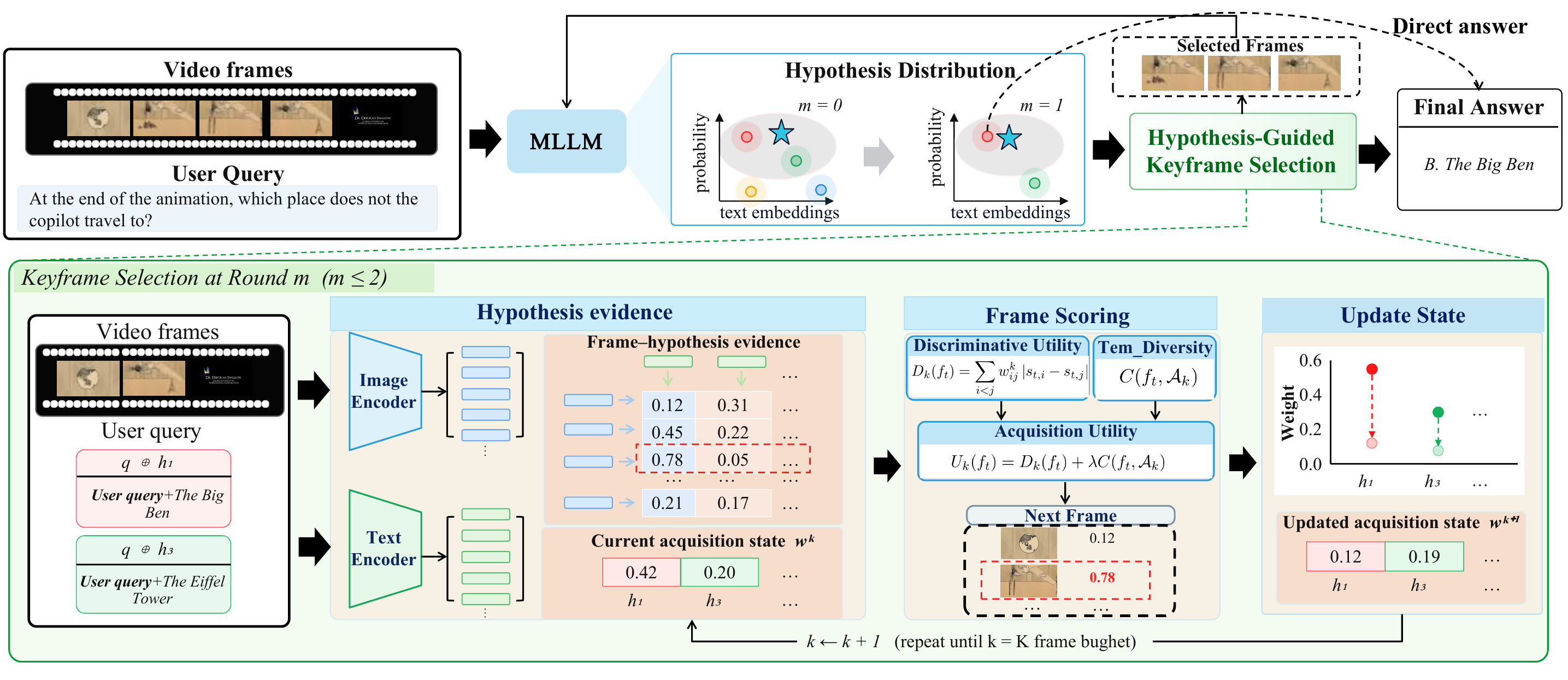}
    \caption{
    \textbf{Overview of HypoKFS.}
    Initial inference from sparse frames constructs a compact hypothesis
    space and determines whether additional computation is required.
    HypoKFS then iteratively selects discriminative frames and updates
    the hypothesis space through renewed inference for at most $M$ rounds.
    }
    \label{fig:method}
\end{figure*}

\subsection{Overview}

Given a long video $V$ and a user query $q$, HypoKFS answers $q$ using a
frozen multimodal large language model (MLLM) under a limited visual
computation budget. We denote the candidate answer hypotheses by
$\mathcal{H}$, containing $C$ alternatives $h_1,\ldots,h_C$.
HypoKFS first performs inference over a small number of uniformly sampled
frames to estimate the answer distribution and construct a compact hypothesis
space. This initial prediction also determines whether further visual
evidence is required. For queries requiring additional evidence, HypoKFS performs iterative
keyframe selection conditioned on the current hypothesis space. Within each
round, candidate frames are evaluated according to their ability to
distinguish the remaining plausible hypotheses, while the acquisition state
is progressively updated to avoid repeatedly emphasizing already covered
hypotheses. The selected frames are then passed to the same frozen MLLM for
renewed inference. The resulting answer distribution updates the hypothesis
space and determines whether another selection round is needed. The process
runs for at most $M$ rounds; if the stopping criterion remains unsatisfied
after the final round, the hypothesis with the highest probability is
returned as the final answer.

\subsection{Initial Hypothesis Space Construction}

We uniformly sample $N$ sparse frames from $V$ and denote this initial frame
set by $\mathcal{F}_{\mathrm{init}}$. Together with the query $q$, these
frames are fed into the frozen MLLM. For each hypothesis $h_i$, we extract
the probability of its first answer token and normalize the probabilities
over all candidate hypotheses:
\begin{equation}
    \mathbf{p}^{0}
    =
    [p_1^0,\ldots,p_C^0],
    \qquad
    p_i^0
    =
    P(h_i\mid\mathcal{F}_{\mathrm{init}},q).
    \label{eq:prior}
\end{equation}
Here, $P(h_i\mid\mathcal{F}_{\mathrm{init}},q)$ denotes the normalized
first-token probability associated with hypothesis $h_i$, rather than the
sequence probability of the complete answer. The distribution
$\mathbf{p}^{0}$ therefore represents the relative support assigned by the
MLLM to different answer hypotheses under the initial sparse visual context.

We sort the hypotheses in descending order of their probabilities and retain
the smallest set whose cumulative probability mass reaches the coverage
threshold $\rho$:
\begin{equation}
    \mathcal{H}_{\rho}
    =
    \{h_{(1)},\ldots,h_{(K_\rho)}\},
    \qquad
    K_\rho
    =
    \min\left\{
        k:
        \sum_{i=1}^{k}p_{(i)}^0\geq\rho
    \right\}.
    \label{eq:hypothesis}
\end{equation}
The resulting $\mathcal{H}_{\rho}$ retains the hypotheses accounting for
most of the current answer probability while excluding low-probability
alternatives from subsequent frame matching. Its size $K_\rho$ is
query-dependent: a concentrated answer distribution preserves fewer
hypotheses, whereas an ambiguous prediction retains more competing
alternatives. After renewed inference in later rounds, the same construction
is applied to the updated first-token distribution.

\paragraph{Selective computation.}
The maximum probability in the current answer distribution is used as the
routing score. If it reaches the routing threshold $\tau$, HypoKFS directly
returns the top-ranked hypothesis and bypasses additional frame matching.
Otherwise, the query enters the keyframe-selection stage. The two thresholds serve different purposes. The routing threshold $\tau$
determines whether additional visual computation is required, whereas the
coverage threshold $\rho$ determines which plausible hypotheses participate
in subsequent matching and discrimination. We use $\rho$ no smaller than
$\tau$, ensuring that queries routed to frame selection retain multiple
competing hypotheses. The same routing criterion is applied after every
renewed inference. If it remains unsatisfied after round $M$, HypoKFS
returns the highest-probability hypothesis from the final answer
distribution.

\subsection{hypothesis-guided Keyframe Selection}

For each routed query, we construct a denser candidate set containing
$N_c$ frames sampled from the video and denote it by $\mathcal{F}$.
Frame selection is conditioned on the hypothesis space
$\mathcal{H}_{\rho}$ obtained from the current inference round. For clarity,
we omit the outer inference-round index in the following formulation.

\paragraph{Frame--hypothesis evidence modeling.}
For each retained hypothesis $h_i$, we concatenate the query $q$ with its corresponding answer hypothesis to form $q\oplus h_i$. A frozen vision--language matching model evaluates the compatibility between candidate frame $f_t$ and hypothesis $h_i$ as
\begin{equation}
    s_{t,i}
    =
    \operatorname{sim}
    \left(
        \phi_v(f_t),
        \phi_t(q\oplus h_i)
    \right),
    \label{eq:similarity}
\end{equation}
where $\phi_v$ and $\phi_t$ denote the visual and textual encoders, respectively, and $\operatorname{sim}$ denotes cosine similarity between their normalized representations. The resulting scores form the frame--hypothesis evidence matrix $\mathbf{S}$ with $N_c$ rows and $K_\rho$ columns, corresponding to candidate frames and retained hypotheses. Each frame--hypothesis similarity is computed when the corresponding hypothesis first participates in selection and is cached thereafter, allowing the same similarities to be reused if the hypothesis remains active in later inference rounds.

\paragraph{Frame scoring.}
At the beginning of each selection round, the current answer probabilities over $\mathcal{H}_{\rho}$ are renormalized to initialize the acquisition weights $w_i^0$, where $w_i^k$ denotes the acquisition importance of hypothesis $h_i$ at selection step $k$. The discriminative utility of candidate frame $f_t$ is defined as
\begin{equation}
    D_k(f_t)
    =
    \sum_{i<j}
    w_{ij}^{k}
    \left|s_{t,i}-s_{t,j}\right|,
    \qquad
    w_{ij}^{k}
    =
    \frac{w_i^{k}w_j^{k}}
    {\sum_{u<v}w_u^{k}w_v^{k}},
    \label{eq:discrimination}
\end{equation}
where $(i,j)$ denotes a retained hypothesis pair and $(u,v)$ enumerates all unordered pairs in $\mathcal{H}_{\rho}$ for normalization. The pair weight $w_{ij}^{k}$ emphasizes competitions involving currently plausible hypotheses, while $|s_{t,i}-s_{t,j}|$ measures how differently frame $f_t$ supports the two hypotheses. To avoid repeatedly selecting temporally nearby evidence, we further combine hypothesis discrimination with temporal diversity:
\begin{equation}
    U_k(f_t)
    =
    \underbrace{
    D_k(f_t)
    }_{\text{hypothesis discrimination}}
    +
    \lambda
    \underbrace{
    C(f_t,\mathcal{A}_k)
    }_{\text{temporal diversity}},
    \qquad
    t_k^{*}
    =
    \arg\max_{t:\,f_t\notin\mathcal{A}_k} U_k(f_t),
    \label{eq:selection}
\end{equation}
where $\mathcal{A}_k$ contains the frames selected before step $k$, $\lambda$ controls the contribution of temporal diversity, and $C(f_t,\mathcal{A}_k)$ is the minimum temporal distance from $f_t$ to the selected frames, normalized by the temporal span of the candidate sequence. The diversity term is omitted when $\mathcal{A}_k$ is empty, and $f_{t_k^*}$ is selected according to the resulting acquisition utility.

\paragraph{Acquisition state update.}
After selecting frame $f_{t_k^*}$, we update the acquisition state as
\begin{equation}
    w_i^{k+1}
    =
    \frac{
        w_i^k \exp(-\alpha s_{t_k^*,i})
    }{
        \sum_j w_j^k \exp(-\alpha s_{t_k^*,j})
    },
    \label{eq:acquisition_update}
\end{equation}
where $s_{t_k^*,i}$ is the similarity between the selected frame and hypothesis $h_i$, $\alpha$ controls the suppression strength, and the denominator normalizes the updated weights over all retained hypotheses. Hypotheses receiving stronger evidence are down-weighted more strongly, reducing the importance of their associated pairs and shifting subsequent selection toward less-covered competitions. The discrimination score is then recomputed using the updated acquisition weights and the cached evidence matrix $\mathbf{S}$, and this \emph{score--select--update} procedure continues until the per-round budget of $K$ keyframes is reached.

\paragraph{Iterative inference and termination.}
The acquisition state $\mathbf{w}^{k}$ is updated within each selection round, whereas the MLLM answer distribution is refreshed only after the acquired keyframes are restored to temporal order and passed back to the frozen MLLM together with the query. The updated distribution reconstructs $\mathcal{H}_{\rho}$ and, when additional evidence is still required, initializes the next round of frame selection. If the updated prediction satisfies the stopping criterion, HypoKFS returns the top-ranked hypothesis; otherwise, the procedure continues for at most $M$ rounds. If the criterion remains unsatisfied after the final round, the hypothesis with the highest probability in the final answer distribution is returned. Neither the MLLM nor the vision--language matching model is updated during inference.

\subsection{Complexity}

Let $r$ denote the fraction of queries routed to keyframe selection and
$\bar{K}_{\rho}$ the average number of distinct retained hypotheses requiring
frame--hypothesis matching across all rounds. For a candidate set of $N_c$
frames, the average matching complexity over all queries is
$\mathcal{O}(rN_c\bar{K}_{\rho})$. Iterative inference does not require
repeated matching for hypotheses that have already appeared, since their
corresponding similarities in $\mathbf{S}$ are cached and reused in later
rounds; only newly introduced hypotheses require additional matching.
Within each round, discrimination scoring, diversity-aware selection, and
acquisition-state updates operate entirely on the cached similarities and
introduce no additional forward passes through the vision--language matching
model. Compared with conventional KFS methods that perform global
image--text matching over the candidate set for every query, HypoKFS
restricts matching to routed queries and their retained hypotheses, thereby
avoiding substantial unnecessary matching computation. The resulting
reduction in frame-matching computation is reported in
Fig.~\ref{fig:belief_analysis}.

\section{Experiments}

\subsection{Experimental Setup}

\paragraph{Datasets and Evaluation.}
We evaluate HypoKFS on three long-video understanding benchmarks:
Video-MME~\cite{fu2025videomme}, LongVideoBench~\cite{wu2024longvideobench},
and MLVU~\cite{zhou2024mlvu}. These benchmarks contain multiple-choice
VideoQA questions covering diverse video durations and reasoning types.
Following the standard evaluation protocols, we report multiple-choice
accuracy on all three benchmarks.

\paragraph{Implementation.}
We evaluate HypoKFS with Qwen3-VL~\cite{bai2025qwen3vl} and
LLaVA-Video~\cite{zhang2024llavavideo}, covering two representative open
video MLLMs and enabling direct comparison with existing training-free
frame-selection methods such as AKS~\cite{tang2025adaptive} and
Q-Frame~\cite{zhang2025qframe}. All MLLMs remain frozen, and HypoKFS
requires no additional training or parameter updates. For the main results,
we consider two per-inference frame budgets, 32 and 64. In each setting, the
number of uniformly sampled frames used for initial inference is kept
identical to the keyframe budget in each subsequent selection round. We allow
at most two rounds of iterative frame selection.Candidate frames are sampled at 1 fps, and CLIP ViT-B/32~\cite{radford2021learning}
serves as the default frozen vision--language encoder for frame--hypothesis matching. Unless otherwise
specified, we set the hypothesis coverage threshold $\rho=0.8$, routing
threshold $\tau=0.8$, suppression coefficient $\alpha=1.0$, and temporal
diversity weight $\lambda=0.1$. All experiments are conducted on NVIDIA
H20-3e GPUs with 141\,GB of memory, using the same hardware configuration
for efficiency and runtime comparisons.

\subsection{Main Results}

\begin{table*}[t]
\centering
\caption{
Performance comparison on LongVideoBench (LVB), MLVU, and Video-MME (V-MME).
For HypoKFS, Frames / Budget reports the average total number of processed
frames and the per-inference frame budget, respectively.
}
\label{tab:main_results}

\small
\renewcommand{\arraystretch}{1.08}
\setlength{\tabcolsep}{5.5pt}

\begin{tabular}{lccccc}
\toprule
Model & Frames & Params & LVB & MLVU & V-MME \\
\midrule

\multicolumn{6}{l}{\textit{Video MLLMs}} \\

GPT-5~\cite{openai2025gpt5}
    & -- & -- & 72.6 & 77.3 & 81.8 \\

Gemini-2.5-Pro~\cite{geminiteam2025gemini25}
    & -- & -- & 77.6 & -- & 84.8 \\

Qwen2.5-VL-72B~\cite{bai2025qwen25vl}
    & 256 & 72B & 60.7 & 74.6 & 73.3 \\

InternVL3-8B~\cite{zhu2025internvl3}
    & 32 & 8B & 62.3 & 71.4 & 66.5 \\

ViLAMP~\cite{cheng2025scaling}
    & 1fps & 7B & 67.5 & -- & 61.2 \\

Keye-VL-1.5~\cite{yang2025keyevl15}
    & 64 & 8B & 73.0 & -- & 66.0 \\

\midrule
\multicolumn{6}{l}{\textit{Training-Free Plug-and-Play Frame Selection}} \\

Qwen3-VL~\cite{bai2025qwen3vl}
    & 128 & 8B & 56.3 & 64.1 & 69.1 \\

\quad + AKS~\cite{tang2025adaptive}
    & 128 & 8B & 58.4 & 65.4 & 69.3 \\

\quad + Q-Frame~\cite{zhang2025qframe}
    & 128 & 8B & 59.1 & 66.6 & 70.3 \\

\quad + FOCUS~\cite{zhu2026focus}
    & 128 & 8B & 59.7 & 67.2 & 70.7 \\

\quad + \textbf{HypoKFS}
    & 82.3/64 & 8B
    & \textbf{60.4} {\scriptsize $\uparrow4.1$}
    & \textbf{70.0} {\scriptsize $\uparrow5.9$}
    & \textbf{72.9} {\scriptsize $\uparrow3.8$} \\

\midrule

LLaVA-Video-7B~\cite{zhang2024llavavideo}
    & 64 & 7B & 58.9 & 71.1 & 64.4 \\

\quad + AKS~\cite{tang2025adaptive}
    & 64 & 7B & 62.5 & 71.2 & 65.1 \\

\quad + Q-Frame~\cite{zhang2025qframe}
    & 64 & 7B & 63.0 & 72.2 & 65.7 \\

\quad + FOCUS~\cite{zhu2026focus}
    & 64 & 7B & 63.5 & 72.7 & 65.4 \\

\quad + \textbf{HypoKFS}
    & 48.2 / 32 & 7B
    & \textbf{64.8} {\scriptsize $\uparrow5.9$}
    & \textbf{77.8} {\scriptsize $\uparrow6.7$}
    & \textbf{67.9} {\scriptsize $\uparrow3.5$} \\

\bottomrule
\end{tabular}
\end{table*}

Table~\ref{tab:main_results} compares HypoKFS with uniform sampling and
existing training-free keyframe selection methods across three long-video
benchmarks. HypoKFS consistently improves both MLLM backbones on all
benchmarks. With Qwen3-VL, HypoKFS improves LongVideoBench from 56.3 to
60.4 ($+4.1$), MLVU from 64.1 to 70.0 ($+5.9$), and Video-MME from 69.1
to 72.9 ($+3.8$). Similar improvements are observed with
LLaVA-Video-7B, where the gains reach $+5.9$, $+6.7$, and $+3.5$ points,
respectively. The consistent improvements across two different MLLM
backbones suggest that the benefit of HypoKFS does not depend on a
particular model architecture or visual input budget.
HypoKFS also compares favorably with existing plug-and-play frame
selectors under the same backbone. On Video-MME with Qwen3-VL, HypoKFS
achieves 72.9, compared with 69.3 for AKS and 70.3 for Q-Frame. Under
LLaVA-Video-7B, HypoKFS reaches 64.8 on LongVideoBench and 67.9 on
Video-MME, consistently improving over the corresponding uniform-sampling
baseline and existing frame selectors. These results indicate that the
advantage does not arise merely from replacing uniform sampling with a
generic relevance-based retrieval stage. Instead, conditioning evidence
acquisition on the hypothesis space produced by initial inference provides
more decision-relevant visual evidence for resolving the remaining answer
ambiguity.

\subsection{Analysis of hypothesis-guided Selective Retrieval}

\begin{figure}[t]
    \centering
    \includegraphics[width=0.325\linewidth]{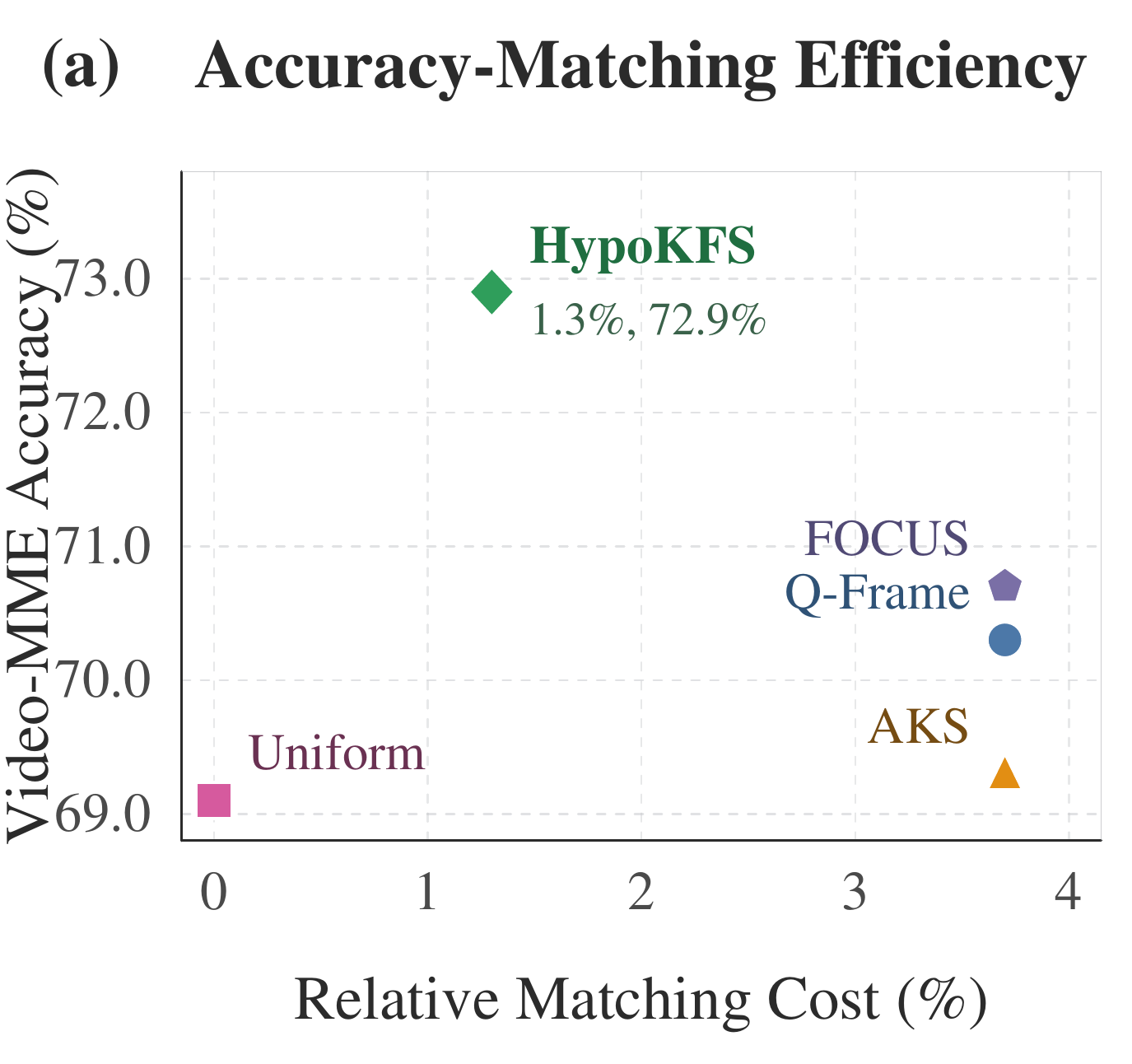}\hfill
    \includegraphics[width=0.325\linewidth]{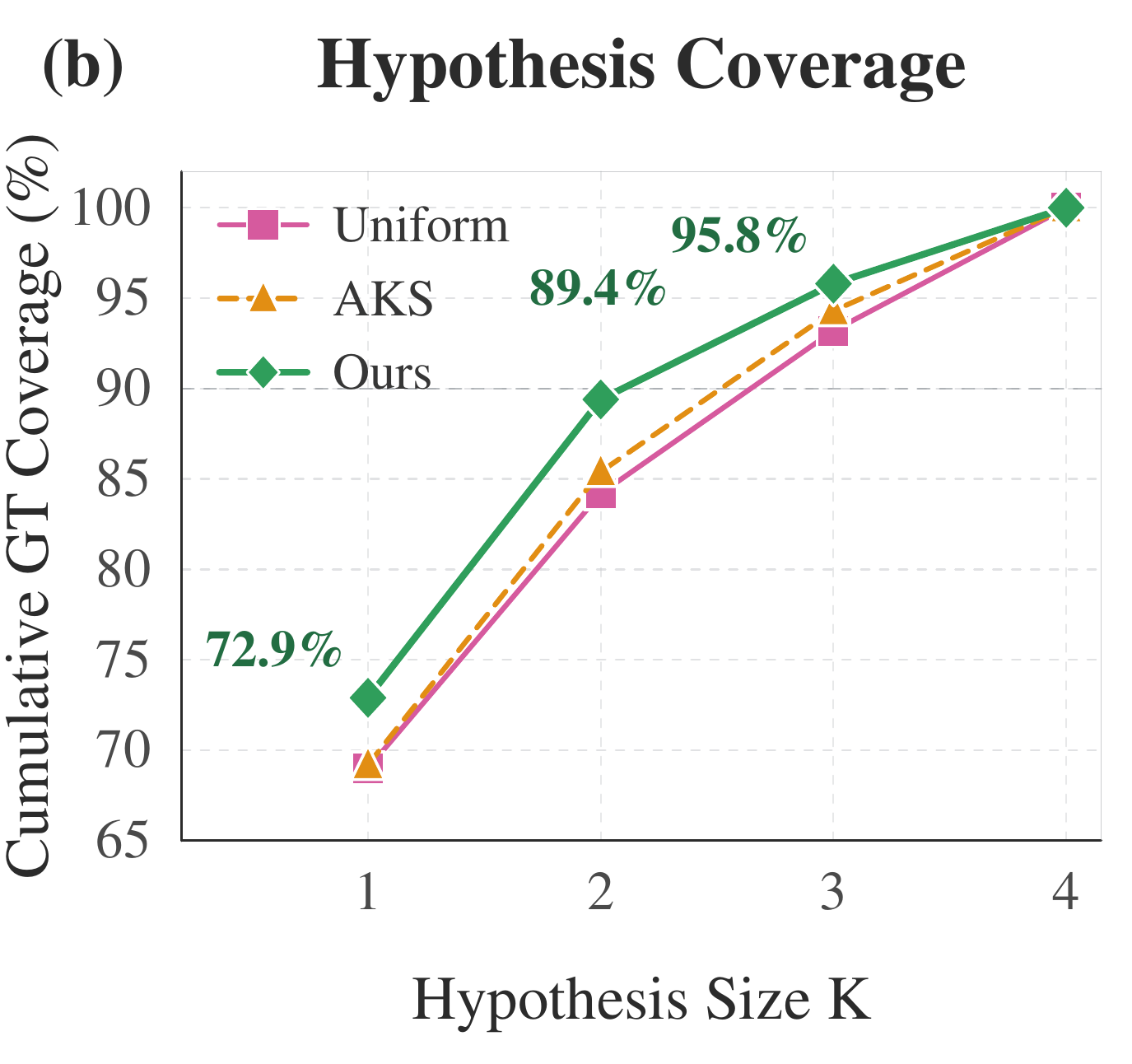}\hfill
    \includegraphics[width=0.325\linewidth]{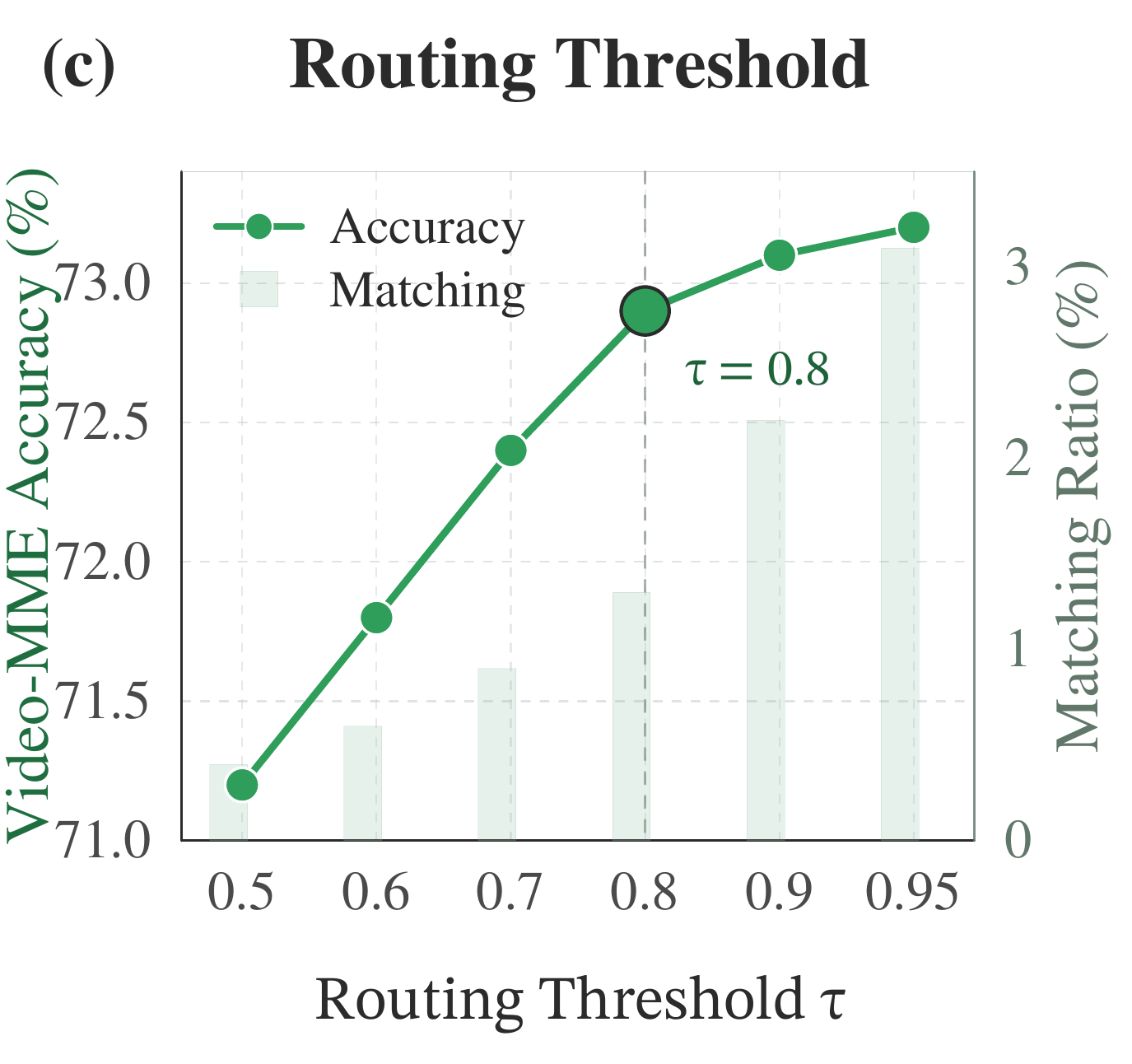}
    \caption{
    Analysis of hypothesis-guided selective retrieval on Video-MME.
    \textbf{(a)} Accuracy--retrieval efficiency.
    \textbf{(b)} Cumulative top-$K$ hypothesis coverage.
    \textbf{(c)} Accuracy--computation trade-off under different thresholds $\tau$.
    }
    \label{fig:belief_analysis}
\end{figure}

Fig.~\ref{fig:belief_analysis} analyzes HypoKFS from three complementary
perspectives: retrieval efficiency, hypothesis concentration, and selective
computation. As shown in Fig.~\ref{fig:belief_analysis}(a), HypoKFS
achieves 72.9\% accuracy with only a 1.3\% matching ratio, providing a
favorable accuracy--computation trade-off compared with conventional
keyframe selectors. This efficiency stems from selective activation:
queries with concentrated initial answer distribution bypass candidate-frame matching,
whereas additional visual computation is reserved for queries that remain
ambiguous.

Fig.~\ref{fig:belief_analysis}(b) examines whether the initial answer distribution provides
a sufficiently informative search space for subsequent evidence acquisition.
The cumulative ground-truth coverage increases rapidly with the number of
retained hypotheses, with the top-2 hypotheses already covering 89.4\% of
correct answers. This observation supports the hypothesis-space construction
in Eq.~\ref{eq:hypothesis}: HypoKFS can focus visual acquisition on a
compact set of plausible alternatives and target evidence that distinguishes
competing hypotheses rather than evidence that is merely relevant to the
query.

Fig.~\ref{fig:belief_analysis}(c) further studies \emph{when} additional
evidence should be acquired by varying the routing threshold $\tau$.
Increasing $\tau$ activates retrieval more aggressively and consequently
raises the matching ratio. Accuracy improves steadily as $\tau$ increases
toward 0.8. Beyond $\tau=0.8$, however, the matching ratio continues to
increase while accuracy improves only marginally, revealing diminishing
returns from additional retrieval. We therefore use $\tau=0.8$ as the
default operating point, achieving 72.9\% accuracy with a 1.3\% matching
ratio.

\subsection{Ablation Studies}

\begin{table}[t]
\centering
\small
\caption{
Ablation studies on Video-MME.
\textbf{(a)} Component ablation across video durations.
\textbf{(b)} Hyperparameter sensitivity to $\alpha$ and $\lambda$ under
keyframe budgets $K=64/128/256$, where each cell reports the corresponding
accuracies in this order.
}
\begin{minipage}[t]{0.41\linewidth}
\vspace{0pt}
\centering
\textbf{(a) Component Ablation}\\[3pt]

\setlength{\tabcolsep}{2.0pt}
\begin{tabular}{ccc|ccc}
\toprule
\multicolumn{3}{c|}{Components}
& \multicolumn{3}{c}{Video-MME} \\
Route & Discrim. & Update
& Short & Medium & Long \\
\midrule
-- & -- & -- & 78.6 & 68.7 & 60.0 \\
\checkmark & -- & -- & 79.5 & 69.5 & 60.8 \\
-- & \checkmark & -- & 80.2 & 70.4 & 61.8 \\
\checkmark & \checkmark & -- & 81.2 & 71.4 & 62.9 \\
\checkmark & \checkmark & \checkmark
& \textbf{82.3} & \textbf{72.5} & \textbf{63.9} \\
\bottomrule
\end{tabular}
\end{minipage}\hfill
\begin{minipage}[t]{0.56\linewidth}
\vspace{0pt}
\centering
\textbf{(b) Hyperparameter Sensitivity}\\[3pt]

\setlength{\tabcolsep}{3.5pt}
\begin{tabular}{c|ccc}
\toprule
\diagbox{$\alpha$}{$\lambda$}
& 0.0 & 0.1 & 0.2 \\
\midrule
0.0
& 72.0/72.3/72.5
& 72.2/72.5/72.7
& 72.0/72.3/72.5 \\
0.5
& 72.3/72.6/72.8
& 72.6/72.9/73.1
& 72.4/72.7/72.9 \\
1.0
& 72.5/72.8/73.0
& \textbf{72.9/73.3/73.5}
& 72.6/73.0/73.2 \\
1.5
& 72.4/72.7/72.9
& 72.7/73.1/73.3
& 72.5/72.8/73.0 \\
2.0
& 72.1/72.4/72.6
& 72.4/72.7/72.9
& 72.2/72.5/72.7 \\
\bottomrule
\end{tabular}
\end{minipage}

\vspace{3pt}
\label{tab:ablation}
\end{table}
Table~\ref{tab:ablation}(a) disentangles the contributions of the three
main components of HypoKFS. Selective routing determines
\emph{when to retrieve} by bypassing candidate-frame matching when the
initial answer distribution is sufficiently concentrated. Introducing routing alone
already improves performance across all video durations, showing that the
initial MLLM inference provides a useful signal for computation allocation.
Hypothesis discrimination instead determines \emph{what to retrieve} by
prioritizing frames that separate plausible hypotheses rather than frames
with generic query relevance. It provides larger gains than routing alone,
particularly on medium and long videos, where the larger temporal search
space makes evidence localization more challenging.
Combining routing and hypothesis discrimination further improves the
results to 81.2, 71.4, and 62.9 on short, medium, and long videos,
respectively, demonstrating that selective computation and
hypothesis-guided evidence acquisition provide complementary benefits.
Finally, incorporating the acquisition-state update yields the full
HypoKFS performance of 82.3, 72.5, and 63.9. By progressively
down-weighting hypotheses already supported by selected evidence, the
update redirects subsequent acquisition toward unresolved competitions
and reduces repeated selection of redundant evidence.

Table~\ref{tab:ablation}(b) studies the sensitivity to the keyframe budget
$K$, suppression coefficient $\alpha$, and temporal diversity weight
$\lambda$. Increasing $K$ from 64 to 128 consistently improves accuracy
across different $(\alpha,\lambda)$ settings, while increasing the budget
further to 256 yields smaller additional gains. This trend indicates diminishing returns beyond $K=64$.
For the acquisition-state update, $\alpha=0$ disables evidence-dependent
suppression and consistently underperforms intermediate values, confirming
the benefit of redirecting acquisition after each selected frame. Meanwhile,
a moderate temporal diversity weight improves performance over
$\lambda=0$, whereas a larger value can overemphasize temporal coverage at
the expense of hypothesis discrimination. 

\subsection{Qualitative Analysis}

\begin{figure}[t]
\centering
\includegraphics[width=\linewidth]{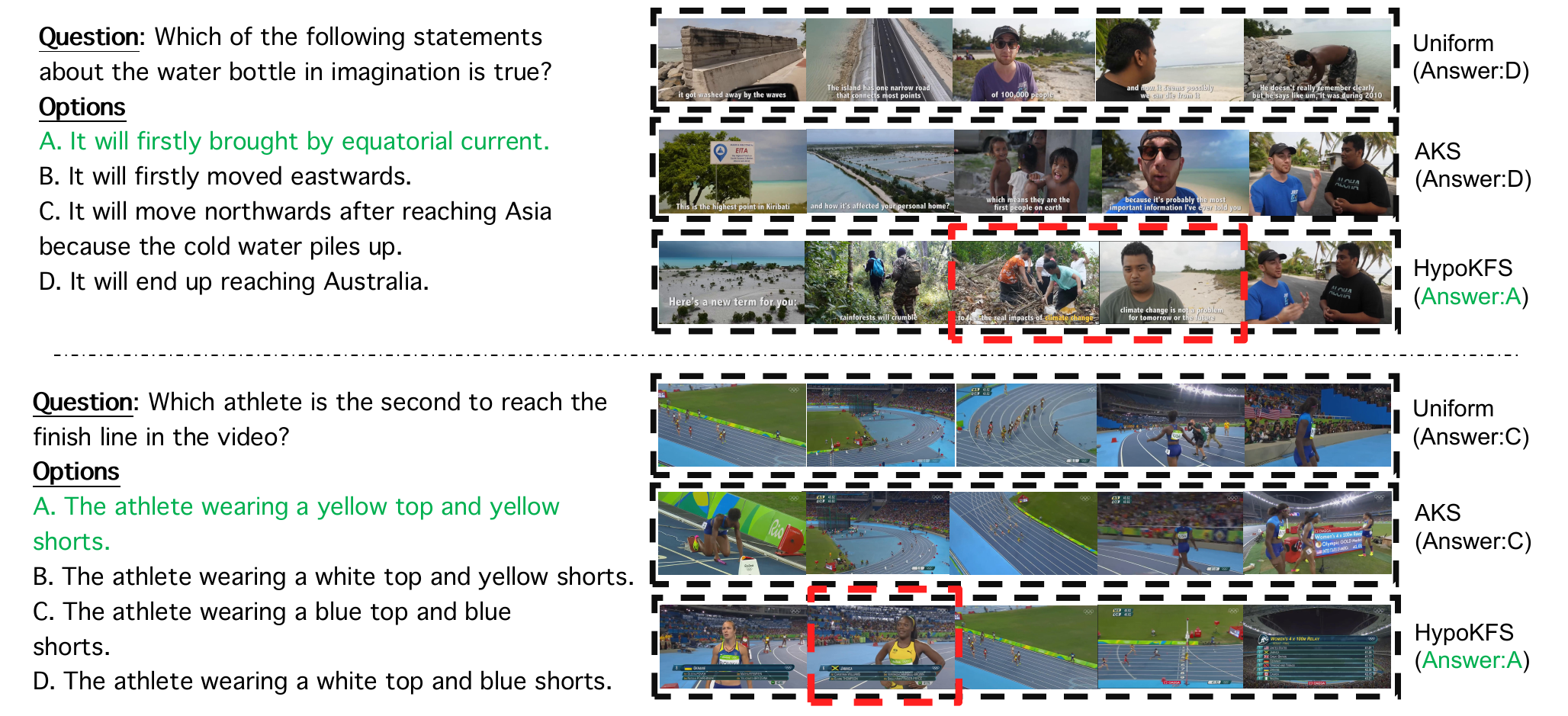}
\caption{\textbf{Qualitative comparison on Video-MME.}
Uniform sampling may miss sparse evidence, while relevance-based retrieval
can favor query-related yet non-discriminative frames. HypoKFS instead
retrieves evidence that resolves competing answer hypotheses.}
\label{fig:case_analysis}
\end{figure}

Fig.~\ref{fig:case_analysis} presents two representative examples from
Video-MME. In the first example, answering the question requires identifying
the visual evidence that distinguishes different descriptions of the water
bottle's trajectory. Uniform sampling and AKS both predict option D, whereas
HypoKFS retrieves frames that better distinguish the competing hypotheses
and correctly selects option A. In the second example, the model must identify
the second athlete reaching the finish line from fine-grained appearance
cues. Uniform sampling and AKS predict option C, while HypoKFS focuses on
frames in which the relevant athlete can be distinguished and correctly
selects option A. These examples illustrate that query relevance alone may
retrieve visually related frames without resolving the answer ambiguity,
whereas HypoKFS explicitly targets evidence that discriminates among the
current hypotheses.

\section{Conclusion}

\paragraph{Summary.}
We presented \textbf{HypoKFS}, a training-free and plug-and-play framework
for hypothesis-guided iterative keyframe selection. HypoKFS first reasons
over sparsely sampled frames to construct a compact hypothesis space, which
determines whether additional visual evidence is required and guides
subsequent frame selection. For ambiguous queries, discriminative frame
scoring and iterative hypothesis refinement focus computation on unresolved
alternatives, while confident predictions bypass unnecessary global frame
matching. Experiments on Video-MME, LongVideoBench, and MLVU demonstrate
consistent accuracy improvements across different MLLM backbones while
substantially reducing frame--hypothesis matching computation.

\paragraph{Limitations and Outlook.}
The current formulation is naturally suited to tasks where a compact set of
answer hypotheses can be reliably constructed, as in multiple-choice VideoQA.
In principle, the same framework can be extended to open-ended settings by
constructing hypotheses from model-generated candidate responses, although
the quality and stability of this hypothesis construction become more
important in such cases. HypoKFS also relies on an additional frozen
vision--language encoder for frame--hypothesis matching, despite substantially
reducing the amount of matching required through selective routing and
cross-round reuse. Finally, subsequent frame selection depends on the quality
of the current hypothesis space, and hypotheses assigned very low probability
in early inference may receive limited attention. Future work may explore
more flexible hypothesis construction and recovery mechanisms for broader
long-video reasoning tasks.

\section{AI use statement}

In this work, we used generative AI tools to assist with literature discovery,
research ideation, implementation discussion, experimental planning, and
writing and editing of the manuscript. Generative AI tools were also used to
assist with mathematical derivations and presentation of the method.
All AI-assisted content, references, methodological suggestions, implementations,
and derivations were independently reviewed and verified by the authors.
All reported experimental results were produced and verified by the authors.
We take full responsibility for the final content of this work.

\bibliography{iclr2027_conference}

@String{Computer = "{IEEE} Computer" }

@article{cheng2025scaling,
  title={Scaling video-language models to 10k frames via hierarchical differential distillation},
  author={Cheng, Chuanqi and Guan, Jian and Wu, Wei and Yan, Rui},
  journal={arXiv preprint arXiv:2504.02438},
  year={2025}
}

@inproceedings{tang2025adaptive,
  title={Adaptive keyframe sampling for long video understanding},
  author={Tang, Xi and Qiu, Jihao and Xie, Lingxi and Tian, Yunjie and Jiao, Jianbin and Ye, Qixiang},
  booktitle={Proceedings of the Computer Vision and Pattern Recognition Conference},
  pages={29118--29128},
  year={2025}
}

@inproceedings{radford2021learning,
  title={Learning transferable visual models from natural language supervision},
  author={Radford, Alec and Kim, Jong Wook and Hallacy, Chris and Ramesh, Aditya and Goh, Gabriel and Agarwal, Sandhini and Sastry, Girish and Askell, Amanda and Mishkin, Pamela and Clark, Jack and others},
  booktitle={International conference on machine learning},
  pages={8748--8763},
  year={2021},
  organization={PmLR}
}

@article{wu2024longvideobench,
  title={Longvideobench: A benchmark for long-context interleaved video-language understanding},
  author={Wu, Haoning and Li, Dongxu and Chen, Bei and Li, Junnan},
  journal={Advances in Neural Information Processing Systems},
  volume={37},
  pages={28828--28857},
  year={2024}
}

@article{2023MovieChat,
  title={MovieChat: From Dense Token to Sparse Memory for Long Video Understanding},
  author={ Song, Enxin  and  Chai, Wenhao  and  Wang, Guanhong  and  Zhang, Yucheng  and  Zhou, Haoyang  and  Wu, Feiyang  and  Guo, Xun  and  Ye, Tianbo  and  Lu, Yang  and  Hwang, Jenq Neng },
  journal={ArXiv},
  volume={abs/2307.16449},
  year={2023},
}

@inproceedings{2025LongVLM,
  title={LongVLM: Efficient Long Video Understanding viaLarge Language Models},
  author={ Weng, Yuetian  and  Han, Mingfei  and  He, Haoyu  and  Chang, Xiaojun  and  Zhuang, Bohan },
  booktitle={European Conference on Computer Vision},
  year={2025},
}

@article{2024LongVILA,
  title={LongVILA: Scaling Long-Context Visual Language Models for Long Videos},
  author={ Chen, Yukang  and  Xue, Fuzhao  and  Li, Dacheng  and  Hu, Qinghao  and  Zhu, Ligeng  and  Li, Xiuyu  and  Fang, Yunhao  and  Tang, Haotian  and  Yang, Shang  and  Liu, Zhijian },
  year={2024},
}

@article{yu2024frame,
  title={Frame-voyager: Learning to query frames for video large language models},
  author={Yu, Sicheng and Jin, Chengkai and Wang, Huanyu and Chen, Zhenghao and Jin, Sheng and Zuo, Zhongrong and Xu, Xiaolei and Sun, Zhenbang and Zhang, Bingni and Wu, Jiawei and others},
  journal={arXiv preprint arXiv:2410.03226},
  year={2024}
}

@inproceedings{liu2025bolt,
  title={Bolt: Boost large vision-language model without training for long-form video understanding},
  author={Liu, Shuming and Zhao, Chen and Xu, Tianqi and Ghanem, Bernard},
  booktitle={Proceedings of the Computer Vision and Pattern Recognition Conference},
  pages={3318--3327},
  year={2025}
}

@article{qian2024streaming,
  title={Streaming long video understanding with large language models},
  author={Qian, Rui and Dong, Xiaoyi and Zhang, Pan and Zang, Yuhang and Ding, Shuangrui and Lin, Dahua and Wang, Jiaqi},
  journal={Advances in Neural Information Processing Systems},
  volume={37},
  pages={119336--119360},
  year={2024}
}

@inproceedings{he2024ma,
  title={Ma-lmm: Memory-augmented large multimodal model for long-term video understanding},
  author={He, Bo and Li, Hengduo and Jang, Young Kyun and Jia, Menglin and Cao, Xuefei and Shah, Ashish and Shrivastava, Abhinav and Lim, Ser-Nam},
  booktitle={Proceedings of the IEEE/CVF conference on computer vision and pattern recognition},
  pages={13504--13514},
  year={2024}
}

@inproceedings{wu2019adaframe,
  title={Adaframe: Adaptive frame selection for fast video recognition},
  author={Wu, Zuxuan and Xiong, Caiming and Ma, Chih-Yao and Socher, Richard and Davis, Larry S},
  booktitle={Proceedings of the IEEE/CVF Conference on Computer Vision and Pattern Recognition},
  pages={1278--1287},
  year={2019}
}

@misc{openai2025gpt5,
  title        = {GPT-5 System Card},
  author       = {{OpenAI}},
  year         = {2025},
  howpublished = {OpenAI}
}

@techreport{geminiteam2025gemini25,
  title       = {Gemini 2.5: Pushing the Frontier with Advanced Reasoning,
                 Multimodality, Long Context, and Next Generation Agentic Capabilities},
  author      = {{Gemini Team, Google}},
  institution = {Google DeepMind},
  year        = {2025}
}

@article{bai2025qwen25vl,
  title   = {Qwen2.5-VL Technical Report},
  author  = {Bai, Shuai and Chen, Keqin and Liu, Xuejing and Wang, Jialin
             and Ge, Wenbin and Song, Sibo and Dang, Kai and Wang, Peng
             and Wang, Shijie and Tang, Jun and others},
  journal = {arXiv preprint arXiv:2502.13923},
  year    = {2025}
}

@article{zhu2025internvl3,
  title   = {InternVL3: Exploring Advanced Training and Test-Time Recipes
             for Open-Source Multimodal Models},
  author  = {Zhu, Jinguo and Wang, Weiyun and Chen, Zhe and Liu, Zhaoyang
             and Ye, Shenglong and Gu, Lixin and Duan, Yuchen and Tian, Hao
             and Su, Weijie and Shao, Jie and others},
  journal = {arXiv preprint arXiv:2504.10479},
  year    = {2025}
}

@article{bai2025qwen3vl,
  title   = {Qwen3-VL Technical Report},
  author  = {Bai, Shuai and others},
  journal = {arXiv preprint},
  year    = {2025}
}

@article{zhang2024llavavideo,
  title   = {Video Instruction Tuning With Synthetic Data},
  author  = {Zhang, Yuanhan and Li, Bo and Liu, Haotian and Lee, Yong Jae
             and Gui, Liangke and Fu, Di and Feng, Jiashi
             and Liu, Ziwei and Li, Chunyuan},
  journal = {arXiv preprint arXiv:2410.02713},
  year    = {2024}
}

@inproceedings{sun2025mdp3,
  title     = {MDP3: A Training-free Approach for List-wise Frame Selection
               in Video-LLMs},
  author    = {Sun, Hui and Lu, Shiyin and Wang, Huanyu and Chen, Qing-Guo
               and Xu, Zhao and Luo, Weihua and Zhang, Kaifu and Li, Ming},
  booktitle = {Proceedings of the IEEE/CVF International Conference
               on Computer Vision},
  pages     = {24090--24101},
  year      = {2025}
}

@inproceedings{zhu2026focus,
  title     = {{FOCUS}: Efficient Keyframe Selection for Long Video Understanding},
  author    = {Zhu, Zirui and Xu, Hailun and Luo, Yang and Liu, Yong
               and Sarkar, Kanchan and Yang, Zhenheng and You, Yang},
  booktitle = {The Fourteenth International Conference on Learning Representations},
  year      = {2026}
}

@inproceedings{zhang2025qframe,
  title     = {Q-Frame: Query-aware Frame Selection and Multi-Resolution
               Adaptation for Video-LLMs},
  author    = {Zhang, Shaojie and Yang, Jiahui and Yin, Jianqin and Luo,
               Zhenbo and Luan, Jian},
  booktitle = {Proceedings of the IEEE/CVF International Conference on
               Computer Vision (ICCV)},
  pages     = {22056--22065},
  year      = {2025}
}

@inproceedings{fu2025videomme,
  title     = {Video-MME: The First-Ever Comprehensive Evaluation Benchmark
               of Multi-modal LLMs in Video Analysis},
  author    = {Fu, Chaoyou and Dai, Yuhan and Luo, Yongdong and Li, Lei
               and Ren, Shuhuai and Zhang, Renrui and Wang, Zihan
               and Zhou, Chenyu and Shen, Yunhang and Zhang, Mengdan
               and others},
  booktitle = {Proceedings of the IEEE/CVF Conference on Computer Vision
               and Pattern Recognition},
  pages     = {24108--24118},
  year      = {2025}
}

@article{yang2025keyevl15,
  title   = {Kwai Keye-VL 1.5 Technical Report},
  author  = {Yang, Biao and Wen, Bin and Ding, Boyang and Liu, Changyi
             and Chu, Chenglong and Song, Chengru and Rao, Chongling
             and Yi, Chuan and Li, Da and Zang, Dunju and others},
  journal = {arXiv preprint arXiv:2509.01563},
  year    = {2025}
}

@article{zhou2024mlvu,
  title   = {MLVU: Benchmarking Multi-task Long Video Understanding},
  author  = {Zhou, Junjie and Shu, Yan and Zhao, Bo and Wu, Boya and
             Xiao, Shitao and Yang, Xi and Xiong, Yongqiang and
             Zhang, Bo and Huang, Tiejun and Liu, Zheng and others},
  journal = {arXiv preprint arXiv:2406.04264},
  year    = {2024}
}

@inproceedings{shen2025longvu,
  title={LongVU: Spatiotemporal Adaptive Compression for Long Video-Language Understanding},
  author={Shen, Xiaoqian and Xiong, Yunyang and Zhao, Changsheng and Wu, Lemeng
  and Chen, Jun and Zhu, Chenchen and Liu, Zechun and Xiao, Fanyi
  and Varadarajan, Balakrishnan and Bordes, Florian and Liu, Zhuang
  and Xu, Hu and Kim, Hyunwoo J. and Soran, Bilge
  and Krishnamoorthi, Raghuraman and Elhoseiny, Mohamed and Chandra, Vikas},
  booktitle={Proceedings of the 42nd International Conference on Machine Learning},
  series={Proceedings of Machine Learning Research},
  volume={267},
  pages={54582--54599},
  year={2025}
}

@inproceedings{tao2025dycoke,
  title={DyCoke: Dynamic Compression of Tokens for Fast Video Large Language Models},
  author={Tao, Keda and Qin, Can and You, Haoxuan and Sui, Yang and Wang, Huan},
  booktitle={Proceedings of the IEEE/CVF Conference on Computer Vision and Pattern Recognition},
  pages={18992--19001},
  year={2025}
}

@inproceedings{yang2025pvc,
  title={PVC: Progressive Visual Token Compression for Unified Image and Video Processing in Large Vision-Language Models},
  author={Yang, Chenyu and Dong, Xuan and Zhu, Xizhou and Su, Weijie
  and Wang, Jiahao and Tian, Hao and Chen, Zhe and Wang, Wenhai
  and Lu, Lewei and Dai, Jifeng},
  booktitle={Proceedings of the IEEE/CVF Conference on Computer Vision and Pattern Recognition},
  pages={24939--24949},
  year={2025}
}

@inproceedings{ye2025voco,
  title={VoCo-LLaMA: Towards Vision Compression with Large Language Models},
  author={Ye, Xubing and Gan, Yukang and Huang, Xiaoke and Ge, Yixiao
  and Tang, Yansong},
  booktitle={Proceedings of the IEEE/CVF Conference on Computer Vision and Pattern Recognition},
  pages={29836--29846},
  year={2025}
}

@inproceedings{hu2025mllmframesel,
  title={M-LLM Based Video Frame Selection for Efficient Video Understanding},
  author={Hu, Kai and Gao, Feng and Nie, Xiaohan and Zhou, Peng
  and Tran, Son and Neiman, Tal and Wang, Lingyun and Shah, Mubarak
  and Hamid, Raffay and Yin, Bing and Chilimbi, Trishul},
  booktitle={Proceedings of the IEEE/CVF Conference on Computer Vision and Pattern Recognition},
  pages={13702--13712},
  year={2025}
}

@inproceedings{li2026maxinfo,
  title={MaxInfo: A Training-Free Key-Frame Selection Method Using Maximum Volume for Enhanced Video Understanding},
  author={Li, Pengyi and Abdullaeva, Irina and Gambashidze, Alexander
  and Kuznetsov, Andrey and Oseledets, Ivan},
  booktitle={Proceedings of the IEEE/CVF Winter Conference on Applications of Computer Vision},
  pages={7198--7207},
  year={2026}
}

@inproceedings{li2026dig,
  title={Divide, then Ground: Adapting Frame Selection to Query Types for Long-Form Video Understanding},
  author={Li, Jialuo and Li, Bin and Li, Jiahao and Lu, Yan},
  booktitle={Proceedings of the IEEE/CVF Conference on Computer Vision and Pattern Recognition},
  pages={11369--11380},
  year={2026}
}

@article{wang2026dafs,
  title={Efficient Frame Selection for Long Videos at Test Time with Attention-Based MLLM Selectors},
  author={Wang, Yilin and Zheng, Xiangxi and Mao, Dongxing and Li, Linjie
  and Yang, Zhengyuan and Yu, Ping and Yan, Rui and Yao, Yuan
  and Wang, Alex Jinpeng},
  journal={arXiv preprint arXiv:2607.15689},
  year={2026}
}
\bibliographystyle{iclr2027_conference}

\appendix

\appendix

\section{Appendix}
\label{sec:appendix}

\subsection{Implementation Details}
\label{sec:appendix_impl}

\paragraph{Multiple-Choice Inference and Hypothesis Construction.}
For multiple-choice VideoQA, we instruct the MLLM to predict the option label
corresponding to the best answer. At the first decoding position, we extract
the logits associated with the candidate answer labels (e.g., A, B, C, and D)
and apply a softmax only over these candidate-label logits. This produces the
answer distribution $\mathbf{p}=[p_1,\ldots,p_C]$ used throughout HypoKFS.
Therefore, $p_i$ denotes the normalized first-token probability of the option
label corresponding to hypothesis $h_i$, rather than the sequence probability
of the complete answer.

The option label and its semantic content are used for different purposes.
The MLLM answer distribution is constructed from the first-token logits of
the option labels, whereas frame--hypothesis matching uses the complete
semantic content of each answer option. Specifically, for hypothesis $h_i$,
we concatenate the question $q$ with the textual content of its corresponding
answer option to form $q\oplus h_i$. The resulting hypothesis-conditioned
text is matched against candidate video frames using the frozen
vision--language encoder. This preserves the standard multiple-choice output
format for MLLM inference while providing semantically informative text for
frame--hypothesis matching.

\paragraph{Video-MME Evaluation Prompt.}
For Video-MME~\cite{fu2025videomme}, we follow the official multiple-choice
evaluation format. The instruction used for inference is:

\begin{quote}
\small
\texttt{Select the best answer to the following multiple-choice question}\\
\texttt{based on the video. Respond with only the letter (A, B, C, or D)}\\
\texttt{of the correct option.}\\
\texttt{Question: [Question]}\\
\texttt{A. [Option A]}\\
\texttt{B. [Option B]}\\
\texttt{C. [Option C]}\\
\texttt{D. [Option D]}\\
\texttt{The best answer is:}
\end{quote}

The same answer format is used for both initial inference and renewed
inference in subsequent rounds, allowing the candidate-label logits to be
extracted consistently throughout the iterative process.

\paragraph{Inference Procedure.}
HypoKFS begins by uniformly sampling a sparse set of frames and performing
an initial forward pass through the frozen MLLM. The first-token logits of
the candidate answer labels are normalized to construct the current answer
distribution. If its maximum probability reaches the routing threshold
$\tau$, the highest-probability hypothesis is returned without additional
frame matching. Otherwise, the smallest probability-ranked hypothesis set
whose cumulative probability reaches $\rho$ is retained for subsequent
keyframe selection.

For a routed query, candidate frames are matched against the retained
hypothesis-conditioned texts. Each frame--hypothesis similarity is computed
when the corresponding hypothesis first participates in selection and is
cached for later reuse. Within each selection round, HypoKFS repeatedly
scores the remaining candidate frames, selects the frame with the largest
acquisition utility, and updates the acquisition state. After the per-round
frame budget is reached, the selected frames are restored to temporal order
and passed to the same frozen MLLM for renewed inference. The updated
first-token answer distribution is then used to reconstruct the hypothesis
space and determine whether another selection round is required. In the main
experiments, we perform at most two keyframe-selection rounds. If the routing
criterion remains unsatisfied after the final round, the hypothesis with the
highest probability in the final answer distribution is returned.

\subsection{Frame-Budget Scaling}
\label{sec:appendix_scaling}

We further examine whether the improvement of HypoKFS persists as the
available visual input increases. We evaluate Qwen3-VL under three frame
settings on Video-MME: 128 uniformly sampled frames, 256 uniformly sampled
frames, and dense sampling at 1 fps. The results are reported in
Table~\ref{tab:frame_scaling}.

\begin{table}[t]
\centering
\small
\setlength{\tabcolsep}{7pt}
\caption{
Scaling performance under different frame settings on Video-MME using
Qwen3-VL.
}
\label{tab:frame_scaling}
\begin{tabular}{lccc}
\toprule
Model & Frames & Params & Video-MME \\
\midrule
Qwen3-VL          & 128   & 8B & 69.1 \\
+ HypoKFS         & 128   & 8B & \textbf{72.9} \\
\midrule
Qwen3-VL          & 256   & 8B & 70.2 \\
+ HypoKFS         & 256   & 8B & \textbf{73.8} \\
\midrule
Qwen3-VL          & 1 fps & 8B & 70.8 \\
+ HypoKFS         & 1 fps & 8B & \textbf{74.3} \\
\bottomrule
\end{tabular}
\end{table}

Increasing the frame density improves the Qwen3-VL baseline from 69.1 with
128 frames to 70.8 at 1 fps, confirming that denser temporal coverage can
recover additional visual information. HypoKFS consistently improves all
three settings, with gains of $+3.8$, $+3.6$, and $+3.5$ percentage points
at 128 frames, 256 frames, and 1 fps, respectively. The improvement remains
stable as the frame density increases, indicating that HypoKFS is not
restricted to low-frame settings. Instead, hypothesis-guided frame selection
remains complementary to increasing the visual input budget by prioritizing
frames that better distinguish the remaining answer hypotheses.

\subsection{Analysis of the Selection Mechanism}
\label{sec:appendix_analysis}

\paragraph{Why Routed Queries Retain Competing Hypotheses.}
The routing threshold $\tau$ and hypothesis-coverage threshold $\rho$ serve
different purposes. A query enters keyframe selection only when its maximum
answer probability is smaller than $\tau$. Since we use $\rho\geq\tau$, a
single hypothesis cannot reach the required cumulative probability mass for
a routed query:
\begin{equation}
    \max_i p_i < \tau \leq \rho.
    \label{eq:appendix_routing}
\end{equation}
It follows that the smallest hypothesis set whose cumulative probability
reaches $\rho$ must contain at least two hypotheses, i.e.,
$K_\rho\geq2$. Pairwise discrimination is therefore well defined whenever
keyframe selection is activated.

\paragraph{Interpretation of Pairwise Competition Weights.}
Recall that the pair weight at selection step $k$ is
\begin{equation}
    w_{ij}^{k}
    =
    \frac{w_i^{k}w_j^{k}}
    {\sum_{u<v}w_u^{k}w_v^{k}}.
    \label{eq:appendix_pair_weight}
\end{equation}
Since the acquisition weights are non-negative, the pair weights are also
non-negative and satisfy $\sum_{i<j}w_{ij}^{k}=1$. The discrimination score
\begin{equation}
    D_k(f_t)
    =
    \sum_{i<j}
    w_{ij}^{k}
    \left|s_{t,i}-s_{t,j}\right|
    \label{eq:appendix_discrimination}
\end{equation}
can consequently be interpreted as a weighted average of the pairwise
separation induced by candidate frame $f_t$. A frame receives a large score
when it provides substantially different support for hypothesis pairs that
currently carry large acquisition weights. Conversely, a frame that assigns
similar scores to competing hypotheses contributes little to resolving the
current ambiguity, even when it is broadly relevant to the question.

\paragraph{Effect of the Acquisition-State Update.}
After selecting frame $f_{t_k^*}$, the acquisition state is updated according
to
\begin{equation}
    w_i^{k+1}
    =
    \frac{
        w_i^k \exp(-\alpha s_{t_k^*,i})
    }{
        \sum_j w_j^k \exp(-\alpha s_{t_k^*,j})
    }.
    \label{eq:appendix_state_update}
\end{equation}
For any two retained hypotheses $h_i$ and $h_j$, their relative acquisition
weight after the update satisfies
\begin{equation}
    \frac{w_i^{k+1}}{w_j^{k+1}}
    =
    \frac{w_i^k}{w_j^k}
    \exp\left(
        -\alpha
        \left[
            s_{t_k^*,i}-s_{t_k^*,j}
        \right]
    \right).
    \label{eq:appendix_relative_update}
\end{equation}
If the selected frame has a larger similarity to $h_i$ than to $h_j$, the
relative acquisition weight assigned to $h_i$ decreases. The update therefore
reduces repeated emphasis on hypotheses that have already received stronger
visual support and redirects subsequent frame selection toward less-covered
hypothesis pairs. Importantly, these acquisition weights are used only within
the frame-selection process and do not modify the MLLM answer distribution.

\paragraph{Cross-Round Similarity Reuse.}
Frame--hypothesis similarities are cached across inference rounds. Once the
similarities between a retained hypothesis and all candidate frames have been
computed, they are reused whenever that hypothesis remains active in a later
round. If renewed inference introduces a previously unseen hypothesis, only
the similarities associated with that hypothesis need to be computed.
Discrimination scoring, temporal-diversity scoring, and acquisition-state
updates operate directly on the cached similarities and therefore require no
additional forward passes through the frozen vision--language matching model.

\end{document}